\documentclass{article} 
\usepackage{iclr2025_conference,times}

\usepackage{amsmath,amsfonts,bm}

\def\eqref#1{equation~\ref{#1}}

\def\1{\bm{1}}

\DeclareMathAlphabet{\mathsfit}{\encodingdefault}{\sfdefault}{m}{sl}
\SetMathAlphabet{\mathsfit}{bold}{\encodingdefault}{\sfdefault}{bx}{n}

\usepackage{hyperref}
\usepackage{url}
\usepackage{graphicx}
\usepackage{multirow}
\usepackage{rotating}
\usepackage[ruled,vlined]{algorithm2e}
\usepackage{amsmath}   
\usepackage{amssymb}
\usepackage{booktabs}    
\usepackage{subcaption}  
\usepackage{caption}     
\usepackage{lipsum}
\usepackage{booktabs} 
\title{Federated Cross-Modal Style-Aware Prompt Generation}

\author{
Suraj Prasad\thanks{Corresponding author: \texttt{suraj.prasad@iitb.ac.in}}, 
Navyansh Mahla, 
Sunny Gupta, 
Amit Sethi \\
Indian Institute of Technology Bombay \\
\texttt{\{suraj.prasad, 210040106, 22d1631, asethi\}@iitb.ac.in}
}

\iclrfinalcopy 
\begin{document}

\maketitle

\begin{abstract}

Prompt learning has propelled vision-language models like CLIP to excel in diverse tasks, making them ideal for federated learning due to computational efficiency. However, conventional approaches that rely solely on final-layer features miss out on rich multi-scale visual cues and domain-specific style variations in decentralized client data. To bridge this gap, we introduce \textbf{FedCSAP (Federated Cross-Modal Style-Aware Prompt Generation)}. Our framework harnesses low, mid, and high-level features from CLIP’s vision encoder alongside client-specific style indicators derived from batch-level statistics. By merging intricate visual details with textual context, FedCSAP produces robust, context-aware prompt tokens that are both distinct and non-redundant, thereby boosting generalization across seen and unseen classes. Operating within a federated learning paradigm, our approach ensures data privacy through local training and global aggregation, adeptly handling non-IID class distributions and diverse domain-specific styles. Comprehensive experiments on multiple image classification datasets confirm that FedCSAP outperforms existing federated prompt learning methods in both accuracy and overall generalization.

\end{abstract}

\section{Introduction}

The integration of VLMs like CLIP into FL settings is hindered by the high computational and communication overhead inherent to training such models in a decentralized environment. This challenge has spurred interest in prompt-based learning techniques that adapt pre-trained models with minimal modifications. In particular, methods like Context Optimization (CoOp) have shown that substituting hand-engineered prompts with learnable soft prompt vectors can efficiently tailor CLIP for specialized downstream tasks within a few-shot learning paradigm \cite{perez2021true,zhou2022learning, zhu2023prompt,yao2023visual}.

Extending these ideas to federated scenarios, Federated Context Optimization (FedCoOp) \cite{guo2023promptfl} adapts the CoOp framework by learning a unified set of prompt vectors across multiple clients, each possessing heterogeneous datasets. While FedCoOp enhances performance on classes encountered during training at individual clients, it still encounters difficulties in generalizing to unseen classes and adapting to tasks with differing contextual nuances (e.g., transitioning from object recognition to texture classification). Moreover, by enabling prompt-based adjustments, these methodologies offer a promising route to mitigate the extensive communication and computational demands typically associated with VLMs in FL environments. In this work, unless stated otherwise, the term ``task'' refers to an image classification dataset.

\begin{figure}
    \centering
    \includegraphics[width=1\linewidth]{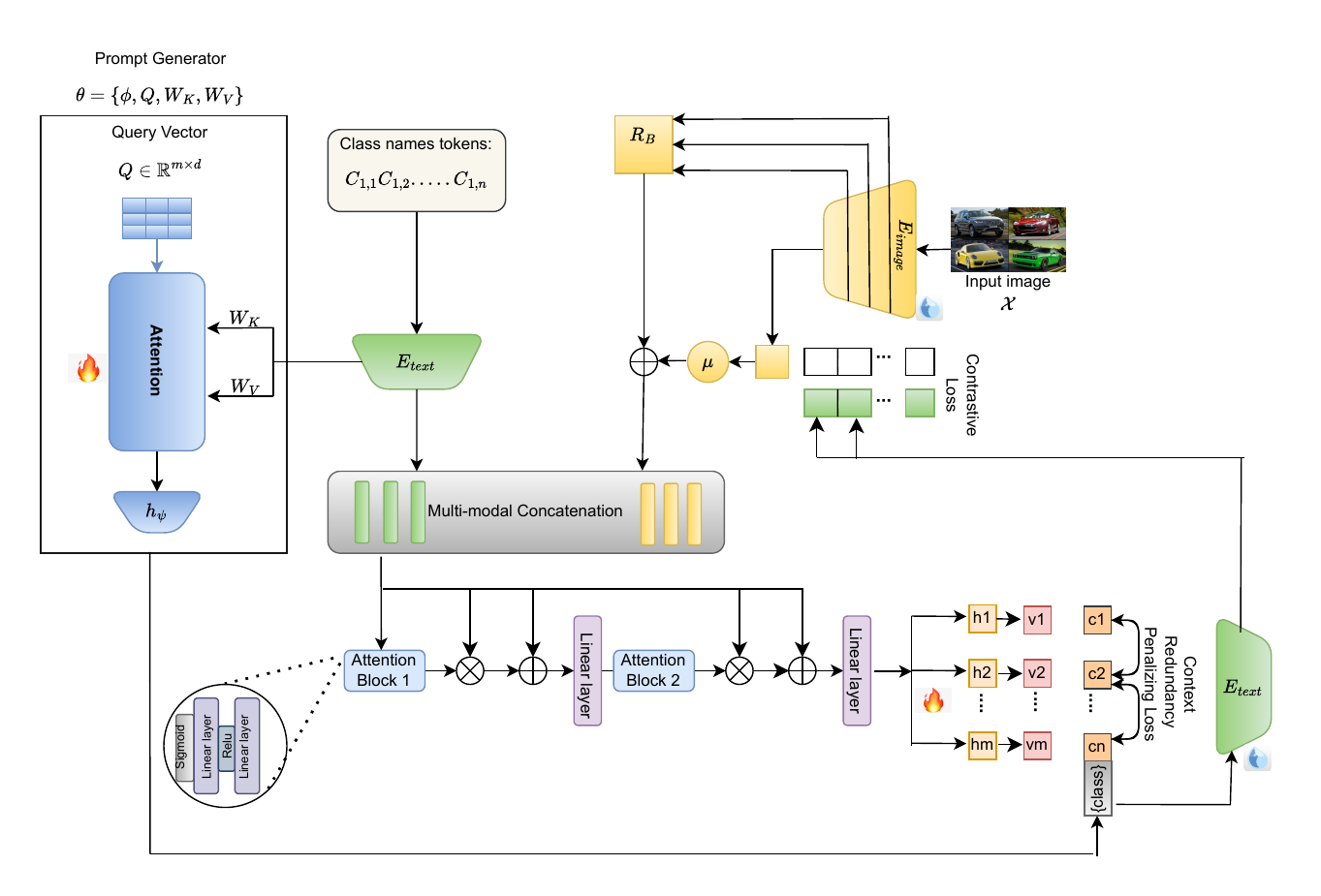}
    \caption{\textbf{FedCSAP:} A federated prompt generator that fuses multi-scale visual features and local style cues with CRP loss to create distinct, context-aware prompt tokens.}
    \label{fig:enter-label}
\end{figure}

Instead of learning a unified set of prompt vectors for all tasks, we convert task-specific text inputs into context-aware prompt vectors.By leveraging the rich semantic details in the text, our method extracts essential contextual information. This enables us to generate prompt vectors that generalize effectively to unseen classification tasks (Fig.~\ref{fig:enter-label}).

Following this idea, we introduce Federated Multi-Scale \& Style-Aware Prompt Generation (FedCMPG), which collaboratively learns a lightweight, unified prompt generator across multiple clients in a federated learning (FL) setting. Unlike existing federated prompt-learning approaches that rely on static learned vectors, our framework dynamically generates multi-modal, style-aware prompt tokens conditioned on both textual and visual information. Each client optimizes the prompt generator locally using its own classification task data, represented by few-shot image-text pairs, while keeping the CLIP encoders frozen. The locally updated parameters are then transmitted to a central FL server, where they are aggregated to form a global prompt generator. An overview of our FedCMPG framework with two remote clients is shown in Fig.~\ref{fig:enter-label}.

At the core of our approach is the Injection Block $ B_{\phi} $, which fuses multi-scale visual features and domain-aware style statistics with class-specific textual embeddings before generating prompt tokens. This block enables our method to enhance task-specific contextualization, ensuring that the learned prompts are both semantically rich and robust to domain shifts. Specifically, our approach includes:

\begin{itemize}
    \item \textbf{Multi-Scale Feature Learning}: We extract low, mid, and high-level features from CLIP’s vision encoder, rather than relying solely on final-layer outputs, to capture fine-grained object textures and global semantics.
    \item \textbf{Style Encoding}: We encode batch-wise statistics (e.g., BN mean values) to represent style variations in different client domains (e.g., lighting conditions, sensor differences), injecting these into the prompt tokens for improved generalization.
    \item \textbf{Context-Aware Prompt Generation}: Instead of using fixed prompts, we concatenate multi-modal information (text embeddings + multi-scale visual/style features) and process them through $ B_{\phi} $, a channel-wise attention mechanism that learns adaptive visual tokens to be merged with prompt embeddings.
    \item \textbf{Federated Optimization}: The prompt generator’s parameters are updated locally and aggregated globally via FedAvg, ensuring an efficient and privacy-preserving FL training pipeline.
\end{itemize}

By training across diverse classification tasks, our prompt generator learns task-specific prompts that enrich CLIP’s representations with domain-aware context information. Our extensive evaluations on nine diverse image classification datasets show that FedCMPG achieves superior generalization, outperforming FedCoOp by 4.32\% on unseen classes and 1.82\% on unseen datasets on average. Furthermore, our efficient parameter-sharing strategy significantly reduces communication overhead, making FedCMPG well-suited for federated environments with limited computational resources.

\section{Related Work}
\paragraph{Visual-Language Model Prompt Learning.} 
Prompt learning, a variant of fine-tuning VLMs, has boosted few-shot performance by optimizing continuous prompt vectors \cite{zhou2022learning} or conditioning them on images \cite{zhou2022conditional}. Subsequent works \cite{khattak2023maple, bulat2023lasp, yao2023visual, chen2022plot, chen2023retrospect, wang2023improving, udandarao2023sus} have further integrated textual and visual cues, soft prompting, and multi-prompt strategies. In contrast, our approach builds a unified model that leverages task-specific text inputs to generalize across diverse tasks and domains.

\paragraph{Federated Learning with Visual-Language Models.} 
Federated Learning (FL) enables decentralized training on heterogeneous data \cite{mcmahan2017communication, li2023fedlga}, and recent works extend VLM fine-tuning to the FL setting \cite{lu2023fedclip, halbe2023hepco, wang2023cooperative, chen2023federated, su2024federated, guo2023promptfl, guo2023pfedprompt}. Unlike methods that rely on fixed client-specific features \cite{yang2023efficient}, our method employs a learnable text-conditioned prompt generator, which improves generalization on both seen and unseen tasks.

\paragraph{Prompt-based Federated Learning.} Prompt learning, initially developed in the field of natural language processing, has expanded its reach to vision language models. Examples include the CLIP model~\cite{radford2021learning}, which originally utilized manually engineered templates. However, more recent advancements have shifted towards developing prompts within a continuous embedding space. Innovations like CoOp~\cite{zhou2022learning} have refined the CLIP model by introducing continuous prompt vectors, sparking a surge of studies aimed at enhancing the efficiency of prompt learning and providing a solid foundation for further investigation~\cite{cui2024harmonizing}. To enhance the integration of global data and address challenges in scenarios with limited user data, FedPrompt~\cite{zhao2023fedprompt} and PromptFL~\cite{guo2023promptfl} have effectively integrated the concept of prompt learning with federated learning~\cite{cai2024fed,cao2023knowledge,li2023fedtp,qiu2024federated,jiangheterogeneous}. To tackle the statistical heterogeneity often found in client data, pFedPrompt~\cite{guo2023pfedprompt} introduces a non-parametric approach, providing each client with a personalized attention module. This module is designed to refine spatial visual features to better align with the unique characteristics of local data. Concurrently, pFedPG~\cite{yang2023efficient} introduces a novel prompt generator located at the server, which customizes prompts for each client, thus enhancing the personalization of the federated learning process. Additionally, a recent study, FedOTP~\cite{li2024global}, leverages optimal transport theory to improve the balance between achieving global consensus and allowing for local customization through a strategic combination of global and local prompts.

\section{Methodology}

Our federated prompt-learning framework is designed to generate context-aware, style-sensitive prompt tokens by fusing multi-scale visual cues and rich textual context. In this section, we first describe the federated problem setup, then detail our multi-scale and style-aware prompt generation module, and finally outline the local training and global aggregation strategy.

\subsection{Problem Setup}

In many federated learning scenarios, clients collect image data from diverse domains. Each client typically possesses a small, non-IID subset of classes, and the images often exhibit distinct style variations (e.g., due to differences in sensors or environmental conditions). Traditional prompt-learning approaches for vision-language models (VLMs) tend to rely solely on final-layer features from a frozen CLIP model, thereby missing valuable low and mid-level visual cues and domain-specific style signals.

Our goal is to design a new architecture that efficiently learns the semantic understanding between prompts and images by:
\begin{itemize}
    \item \textbf{Capturing Multi-Scale Visual Cues:} Extracting features from multiple layers of the CLIP vision encoder to retain fine-grained details as well as global semantics.
    \item \textbf{Encoding Domain-Specific Style:} Incorporating batch-level style statistics (i.e. batch-normalization) to adapt the prompts to local client variations.
    \item \textbf{Enforcing Token Diversity:} Using a Context Redundancy Penalizing (CRP) ~\cite{crp_loss_2023} loss to ensure that the generated prompt tokens are distinct and non-redundant.
\end{itemize}

\subsection{Multi-Scale \& Style-Aware Prompt Generation}

At the heart of our approach is a prompt generation module \( f_\theta \) that transforms task-specific text inputs into adaptive prompt vectors, and an injection block \( B_\phi \) that fuses multi-scale visual features with domain-aware style cues as shown in the Fig.~\ref{fig:enter-label}

\subsubsection{Textual Context via Cross-Attention}

Candidate class names are first embedded into a rich semantic space using a frozen CLIP text encoder:
\begin{align}
\mathcal{T} &= \{ E_{\text{text}}(c_j) \}_{j=1}^{n} \in \mathbb{R}^{n \times d}.
\end{align}

Instead of using fixed prompt vectors, our prompt generator employs a lightweight cross-attention mechanism to dynamically extract context from these text embeddings. This mechanism leverages a learnable query matrix \( Q \) of shape \( (m \times d) \) (with \( m = 4 \) prompt tokens and \( d = 512 \) matching the CLIP text encoder’s output) that is randomly initialized. The learnable queries enable the model to “query” the transformed text embeddings:

\begin{equation}
K_{\mathcal{T}} = \mathcal{T} \, W_K, \quad V_{\mathcal{T}} = \mathcal{T} \, W_V.
\end{equation}

where $W_K$ and $W_V$ are projection matrices. The hidden layer $h_\phi$ projects cross-attention layer output to prompt vectors $\mathcal{P}$. The prompt generator is then defined as:

\begin{equation}
\mathcal{P} = f_\theta(\mathcal{T}) = h_\phi\Big( \text{CrossAttention}(Q, K_{\mathcal{T}}, V_{\mathcal{T}}) \Big)
\end{equation}

This cross-attention mechanism allows our model to dynamically fuse the rich, context-specific information from the text embeddings into the prompt vectors, making them adaptive to each task's unique semantics.

\subsubsection{Multi-Scale Visual and Style Feature Extraction}

To capture visual details at different levels, we extract features from multiple layers as shown in Fig.~\ref{fig:enter-label} \( l \) of CLIP’s vision encoder:

\begin{equation}
\mathbf{f}_v^l(x) \in \mathbb{R}^{W_l \times H_l \times C_l}
\end{equation}

Each feature map is globally average pooled (GAP) ~\cite{lin2013network} to yield a compact representation:

\begin{equation}
\widehat{\mathbf{f}}_v^l(x) = \mathrm{GAP}\bigl(\mathbf{f}_v^l(x)\bigr) \in \mathbb{R}^{C_l}
\end{equation}

We concatenate these \( L \) pooled vectors to form a multi-scale content vector: \( \widehat{\mathbf{F}}(x) \).

To account for domain-specific variations, we compute style cues from batch-level statistics (e.g., the batch-normalization mean \( \boldsymbol{\mu}_i \)) and append these to the visual features:
\[
\mathbf{F}(x) = \Bigl[ \widehat{\mathbf{F}}(x); \boldsymbol{\mu}_i \Bigr].
\]
We then form a multi-modal embedding by concatenating the text features as shown in Fig.~\ref{fig:enter-label} with the style cues:
\[
\mathbf{M}(x) = \Bigl[ \mathcal{T}; \mathbf{F}(x) \Bigr].
\]

\subsubsection{Injection Block for Feature Fusion}

Inspired by Squeeze-and-Excitation (SE-Net)~\cite{hu2018senet}, Injection Block \( B_\phi \) ~\cite{crp_loss_2023} integrates the visual and style features with the text context through sequential channel-wise attention layers. For \( Q \) such layers, the operations are defined as:
\begin{equation}
    \mathbf{O}_1 = \mathbf{F}(x) \otimes A_1(\mathbf{F}(x)) + \mathbf{F}(x)
\end{equation}
\begin{equation}    
\mathbf{O}_q = \mathbf{O}_{q-1} \otimes A_q(\mathbf{O}_{q-1}) + \mathbf{O}_{q-1},\quad q = 2,\dots, Q,
\end{equation}

where \( \otimes \) denotes element-wise (channel-wise) multiplication, and each attention module is given by:
\[
A_q(\mathbf{v}) = \sigma\Big( W_2 \, \delta(W_1 \mathbf{v}) \Big),
\]

Here \( \delta \) represents the ReLU function, \( \sigma \) denotes the Sigmoid, and \( W_1, W_2 \) are learnable parameters optimized during training.

After these attention blocks, the final feature \( \mathbf{O}_Q \) is projected through linear layers \( \{h_m\} \) to produce \( M \) visual tokens:
\[
v_m = h_m(\mathbf{O}_Q), \quad m = 1, \dots, M.
\]
The learned prompt tokens \( \{ c_m \} \) from \( f_\theta \) are then enhanced by injecting the visual cues:
\[
c'_m = c_m + v_m, \quad m = 1, \dots, M.
\]
Finally, for each candidate class, the augmented prompt vector is formed as:
\[
t_j = \Bigl\{ [v_1 + c_1], [v_2 + c_2], \dots, [v_M + c_M], [CLS_y] \Bigr\}.
\]

The classification probability is computed using the cosine similarity between the CLIP image features and these prompt tokens (scaled by a temperature \( \tau \)). To train the model, we optimize a combined loss:
\begin{equation}
L_{\text{total}} = L_{\text{ce}} + \lambda \, L_{\text{CRP}}    
\end{equation}

where \( L_{\text{ce}} \) is the cross-entropy loss which is the negative log-likelihood of the correct label:
     \[
     L_i(\theta^r, \phi^r; \mathcal{T}_i) = - \mathbb{E}_{(x,y) \sim \mathcal{D}_i}\Bigl[y \log p_{\theta^r, \phi^r}(y \mid x, \mathcal{T}_i)\Bigr],
     \]
     where \( p_{\theta^r, \phi^r}(y \mid x, \mathcal{T}_i) \) is computed based on the cosine-similarity between the CLIP image features and the generated prompt tokens:
\begin{equation}
p_{\theta,\phi} (y = j | x, \mathcal{T}) = \frac{\exp(\cos (E_{image}(x), E_{text}(t_j)) / \tau)}
{\sum_{i}^{n} \exp(\cos (E_{image}(x), E_{text}(t_i)) / \tau)}
\end{equation}
where as the CRP loss is shown as:

\begin{equation}
L_{\text{CRP}} = \sum_{j \neq l} \Bigl| c'_j \cdot c'_l - I \Bigr|\end{equation}

encourages low pairwise similarity among prompt tokens, where 
I represents the  Identity Matrix.

\subsection{Local Training and Server Aggregation}
\label{sec:local-training}

Our framework is embedded within a federated learning paradigm that preserves data privacy and minimizes communication overhead. We outline the training pipeline
of our FedTPG in Algorithm ~\ref{alg:ourmodel}  The key steps are as follows:

\begin{enumerate}
    \item \textbf{Model Distribution:}  
    In each communication round \( r \), the central server sends the current global parameters \( \theta^r \) (for \( f_\theta \)) and \( \phi^r \) (for \( B_\phi \)) to a randomly selected subset of clients.

    \item \textbf{Local Training:}  
    Each client \( i \) uses its local dataset \( \mathcal{D}_i \) and frozen CLIP encoders to extract multi-scale features and style cues. The local prompt generator \( f_\theta \) produces task-specific prompt tokens from the text embeddings, and the Injection Block fuses these with the visual/style features. Clients compute the classification loss \( L_{\text{ce}} \) and the CRP loss \( L_{\text{CRP}} \) and update their local parameters \( \theta_i \) and \( \phi_i \) via a few steps of gradient descent (e.g., using SGD).

     Using an optimizer (such as SGD) and a local learning rate \( \eta_r \), each client performs \( K \) gradient steps to update its local parameters:
   \[
   \theta_{i}^{r+1} = \text{SGD}_K\bigl(\eta_r, \theta^r, \mathcal{T}_i, L_i, L_{\text{CRP}}\bigr),
   \]
   \[
   \phi_{i}^{r+1} = \text{SGD}_K\bigl(\eta_r, \phi^r, \mathcal{T}_i, L_i, L_{\text{CRP}}\bigr).
   \]

    \item \textbf{Server Aggregation:}  
    After local updates, each client transmits its updated parameters back to the server. The server aggregates these updates using a standard federated averaging (FedAvg) scheme:
    \[
    \theta^{r+1} = \frac{1}{|S_r|} \sum_{i \in S_r} \theta_i^{r+1}, \quad \phi^{r+1} = \frac{1}{|S_r|} \sum_{i \in S_r} \phi_i^{r+1}.
    \]
    This iterative aggregation across \( R \) rounds produces a global prompt generator that adapts to diverse client data without sharing raw images or local style references.
\end{enumerate}

By integrating multi-scale visual information, local style cues, and dynamic text-context fusion through cross-attention, our approach generates robust and diverse prompt tokens. This enables improved generalization across both seen and unseen classes in heterogeneous federated environments.


\begin{algorithm}[ht]
\caption{OurModel Algorithm}
\label{alg:ourmodel}
\KwIn{%
  No.\ of communication rounds $R$, 
  no.\ of local epochs $K$, 
  initialization parameters $\theta^0$ and $\phi^0$.%
}

\textbf{Server executes:}\\[5pt]
Initialize prompt generator $f_{\theta}$ with parameters $\theta^0$, 
and initialize $B(\phi)$ with $\phi^0$.\\
\For{$r \leftarrow 0$ \KwTo $R$}{
    Pick a random subset of remote clients as $S^r$.\\
    \For{$i \in S^r$ in parallel}{
        Send the current global model $\theta^r$ and $\phi^r$ to client $i$.\\
        Receive locally updated $\theta_i^{r+1}$ and $\phi_i^{r+1}$ 
        from \emph{Local Client Training}.\\
    }
    Aggregate the updated model parameters:\\
    \[
       \theta^{r+1} \;=\; \frac{1}{\lvert S^r\rvert} \sum_{i \,\in\, S^r} \theta_i^{r+1},
       \quad
       \phi^{r+1} \;=\; \frac{1}{\lvert S^r\rvert} \sum_{i \,\in\, S^r} \phi_i^{r+1}.
    \]
}
Obtain the final model parameters $\theta^R$ and $\phi^R$.\\[5pt]

\textbf{Local Client Training:}\\[3pt]
Obtain the set of class name embeddings 
$\mathcal{T}_i = \{\mathcal{E}_{\text{text}}(c_{i,j})\}_{j=1}^{n_i}$.\\[3pt]

Obtain the set of encoding style and multi-scale content features:
\[
  \hat{F}(x) 
    \;=\; 
    \bigl[\hat{f}_{v}^{1}(x);\;\dots;\;\hat{f}_{v}^{L}(x)\bigr],
  \qquad
  F(x) 
    \;=\; 
    \bigl[\hat{F}(x_i);\;\mu_i\bigr].
\]
Fusion of visual and text features:
\[
  H(x) \;=\; \bigl[F(x);\;\mathcal{T}_i\bigr].
\]

\For{$k \leftarrow 0$ \KwTo $K$}{
  Generate the context prompt vectors 
  \[
    P_i^r = f_{\theta_i^r}(\mathcal{T}_i).
  \]
  Generate the multi-modal fusion prompt vector:
  \[
    P_{M_i}^r 
      = B_{\phi_i^r}(H).
  \]
  Get the prompt vectors for each class:
  \[
    t_{i,j}^r 
      = \bigl(P_i^r + P_{M_i}^r\bigr) 
        \;\cup\; 
        \{\,c_{i,j}\}.
  \]
  Update parameters 
  \(
    \theta^r \to \theta_i^{r+1}
  \)
  and 
  \(
    \phi^r \to \phi_i^{r+1}
  \)
  locally on $(x,y)\sim D_i$.\\
}
\end{algorithm}
\section{Experiments}

\subsection{Experiment Setup}
\textbf{Baselines:} We compare FedCSAP with: (i) CLIP using a hand-crafted prompt (e.g., ``a photo of a [class]''), (ii) locally-trained CoOp~\cite{zhou2022learning}, (iii) FedCoOp~\cite{guo2023promptfl} using federated averaging, (iv) FedKgCoOp~\cite{yao2023visual} with prompt regularization, (v) FedCoCoOp~\cite{zhou2022conditional} that conditions prompt generation on images, and (vi) FedMaple~\cite{khattak2023maple} learning prompts for both vision and text encoders. All federated variants employ FedAvg~\cite{mcmahan2017communication} to learn a unified model across clients.

\textbf{Implementation:} All methods use a frozen CLIP ViT-B/16 backbone. FedCSAP's prompt generator consists of a four-head cross-attention module with layer norm and a two-layer MLP (\(h_{\phi}\)), with all vector dimensions set to 512. The prompt vectors have a length \(m=4\) and dimension \(d=512\), and the model includes an Injection Block \(B_{\phi}\).

\textbf{Datasets:} We employ nine image datasets including \textit{Caltech101} \cite{fei2004learning}, \textit{OxfordPets} \cite{parkhi2012cats}, \textit{StanfordCars} \cite{krause20133d}, \textit{Flowers102} \cite{nilsback2008automated}, \textit{Food101} \cite{bossard2014food}, \textit{FGVCAircraft} \cite{maji2013fine}, \textit{SUN397} \cite{xiao2010sun}, \textit{UCF101} \cite{soomro2012ucf101}, and \textit{DTD} \cite{cimpoi2014describing}.

 We divide the classes in each dataset into two equal groups: one set serves as base classes, and the other as new classes. The images from base classes are used for training, while those from new classes help evaluate how well the model generalizes. We conduct all experiments in a non-IID federated learning (FL) setting, where the base classes from nine datasets are distributed across multiple clients. Each client is assigned 20 completely distinct classes, with each class containing eight labeled images for few-shot training. We measure classification accuracy across three levels: individual client tasks, base classes (aggregating classes from multiple clients), and new classes. The overall performance is represented using the harmonic mean (HM) of these three accuracy scores, and all results are averaged over three independent runs.

\begin{table}[!htbp]
    \centering
    \begin{subtable}[t]{0.48\linewidth}
        \centering
        \caption{Averaged results over 9 datasets}
        \label{tab:averaged_results}
        \begin{tabular}{lcccc}
        \toprule
        Method      & Local & Base  & New   & HM    \\
        \midrule
        CLIP        & 76.79 & 70.40 & 75.08 & 73.98 \\
        CoOp        & \textbf{83.41} & 72.05 & 71.27 & 75.09 \\
        FedCoOp     & 80.12 & 74.42 & 71.62 & 75.22 \\
        FedKgCoOp   & 79.56 & 72.34 & 74.18 & 75.22 \\
        FedCoCoOp   & 81.36 & 72.85 & 67.74 & 73.54 \\
        FedMaple    & 81.05 & \textbf{74.50} & 70.40 & 73.24 \\
        \bottomrule
        FedCSAP     & 79.40 & 73.29 & \textbf{75.61} & \textbf{76.06} \\
        \bottomrule
    \end{tabular}
    \end{subtable}
    \hfill
    \begin{subtable}[t]{0.48\linewidth}
        \centering
        \caption{Caltech101}
        \label{tab:caltech101}
        \begin{tabular}{lcccc}
            \toprule
            Method      & Local  & Base   & New    & HM     \\
            \midrule
            CLIP        & 97.37  & 96.56  & 94.37  & 96.10\\
            CoOp        & \textbf{98.10}  & 93.83  & 95.10  & 94.56  \\
            FedCoOp     & 97.04  & \textbf{97.15}  & 93.23  & 95.77  \\
            FedKgCoOp   & 97.27  & 97.01  & 94.47  & \textbf{96.23}  \\
            FedCoCoOp   & 96.46  & 93.91  & 91.74  & 94.00  \\
            FedMaple    & 97.08  & 95.28  & 89.72  & 93.92  \\
            \bottomrule
            FedCSAP        & 96.91  & 96.32  & \textbf{95.31}  & 96.17  \\
            \bottomrule
        \end{tabular}
    \end{subtable}
    
    \vspace{1em}
    
    \begin{subtable}[t]{0.48\linewidth}
        \centering
        \caption{Flowers102}
        \label{tab:flowers102}
        \begin{tabular}{lcccc}
            \toprule
            Method      & Local  & Base   & New    & HM     \\
            \midrule
            CLIP        & 82.52  & 72.02  & 76.93  & 76.91  \\
            CoOp        & \textbf{97.27}  & 69.84  & 72.15  & 78.16  \\
            FedCoOp     & 84.81  & 76.05  & 70.12  & 79.07  \\
            FedKgCoOp   & 95.32  & 74.12  & 70.23  & 79.25  \\
            FedCoCoOp   & 94.44  & 71.73  & 68.34  & 78.07  \\
            FedMaple    & 90.76  & \textbf{76.44}  & 68.51  & 78.35 \\
            \bottomrule
            FedCSAP        & 83.51  & 71.04  & \textbf{77.24}  & \textbf{76.93}  \\
            \bottomrule
        \end{tabular}
    \end{subtable}
    \hfill
    \begin{subtable}[t]{0.48\linewidth}
        \centering
        \caption{FGVC Aircraft}
        \label{tab:fgvc_aircraft}
        \begin{tabular}{lcccc}
            \toprule
            Method      & Local  & Base   & New    & HM     \\
            \midrule
            CLIP        & 30.75  & 27.63  & 30.85  & 29.67  \\
            CoOp        & \textbf{36.57}  & 28.59  & 28.27  & 30.71  \\
            FedCoOp     & 35.92  & \textbf{32.82}  & 30.89  & 33.08  \\
            FedKgCoOp   & 34.55  & 31.18  & 32.13  & 32.56  \\
            FedCoCoOp   & 35.21  & 31.93  & 22.67  & 28.89  \\
            FedMaple    & 35.09  & 31.42  & 31.94  & 32.74  \\
            \bottomrule
            FedCSAP        & 35.29  & 32.71  & \textbf{32.20}  & \textbf{33.35}  \\
            \bottomrule
        \end{tabular}
    \end{subtable}
    
    \vspace{1em}
    
    \begin{subtable}[t]{0.48\linewidth}
        \centering
        \caption{CF101}
        \label{tab:ucf101}
        \begin{tabular}{lcccc}
            \toprule
            Method      & Local  & Base   & New    & HM     \\
            \midrule
            CLIP        & 80.77  & 70.18  & \textbf{76.78}  & 75.65  \\
            CoOp        & \textbf{87.93}  & 70.06  & 68.50  & 74.54  \\
            FedCoOp     & 86.17  & \textbf{75.63}  & 70.51  & 76.90  \\
            FedKgCoOp   & 83.48  & 73.52  & 74.13  & 76.78  \\
            FedCoCoOp   & 83.87  & 75.32  & 68.25  & 75.28  \\
            FedMaple    & 84.04  & 75.01  & 66.45  & 74.48  \\
            \bottomrule
            FedCSAP        & 84.43  & 73.84  & 75.72  & \textbf{77.73}  \\
            \bottomrule
        \end{tabular}
    \end{subtable}
    \hfill
    \begin{subtable}[t]{0.48\linewidth}
        \centering
        \caption{OxfordPets}
        \label{tab:oxford_pets}
        \begin{tabular}{lcccc}
            \toprule
            Method      & Local  & Base   & New    & HM     \\
            \midrule
            CLIP        & 91.39  & 91.36  & \textbf{97.09}  & 93.19  \\
            CoOp        & 94.36  & 94.88  & 96.14  & 95.12  \\
            FedCoOp     & 93.75  & 94.11  & 96.33  & 94.72  \\
            FedKgCoOp   & 92.48  & 94.40  & 96.89  & 94.56  \\
            FedCoCoOp   & 92.30  & 94.77  & 96.30  & 94.43  \\
            FedMaple    & 94.52  & \textbf{95.44}  & 96.84  & \textbf{95.59}  \\
            \bottomrule
            FedCSAP        & \textbf{95.06}  & 95.01  & 95.70  & 95.26  \\
            \bottomrule
        \end{tabular}
    \end{subtable}
    
    \vspace{1em}
    
    \begin{subtable}[t]{0.48\linewidth}
        \centering
        \caption{Food101}
        \label{tab:food101}
        \begin{tabular}{lcccc}
            \toprule
            Method      & Local  & Base   & New    & HM     \\
            \midrule
            CLIP        & \textbf{94.65}  & \textbf{90.23}  & 90.80  & \textbf{91.85}  \\
            CoOp        & 93.60  & 88.30  & 88.22  & 89.97  \\
            FedCoOp     & 93.46  & 88.59  & 88.58  & 90.15  \\
            FedKgCoOp   & 93.66  & 87.96  & 89.15  & 90.19  \\
            FedCoCoOp   & 93.53  & 87.08  & 84.87  & 88.34  \\
            FedMaple    & 93.95  & 89.43  & 89.60  & 90.95 \\
            \bottomrule
            FedCSAP        & 93.80  & 89.36  & \textbf{91.06}  & 91.37  \\
            \bottomrule
        \end{tabular}
    \end{subtable}
    \hfill
    \begin{subtable}[t]{0.48\linewidth}
        \centering
        \caption{DTD}
        \label{tab:dtd}
        \begin{tabular}{lcccc}
            \toprule
            Method      & Local  & Base   & New    & HM     \\
            \midrule
            CLIP        & 53.62  & 53.11  & 58.44  & 54.94  \\
            CoOp        & \textbf{72.59}  & \textbf{73.24}  & 55.36  & \textbf{65.98}  \\
            FedCoOp     & 69.12  & 68.88  & 52.32  & 62.36  \\
            FedKgCoOp   & 58.53  & 58.63  & 59.46  & 58.87  \\
            FedCoCoOp   & 68.33  & 68.63  & 45.73  & 58.71  \\
            FedMaple    & 68.17  & 68.51  & 47.14  & 59.41  \\
            \bottomrule
            FedCSAP        & 61.11  & 61.11  & \textbf{61.96}  & 61.39  \\
            \bottomrule
        \end{tabular}
    \end{subtable}
    
    \vspace{1em}
    
    \begin{subtable}[t]{0.48\linewidth}
        \centering
        \caption{StanfordCars}
        \label{tab:stanford_cars}
        \begin{tabular}{lcccc}
            \toprule
            Method      & Local  & Base   & New    & HM     \\
            \midrule
            CLIP        & 71.25  & 63.24  & \textbf{74.98}  & 69.43  \\
            CoOp        & \textbf{78.73}  & 62.99  & 70.30  & 70.0  \\
            FedCoOp     & 71.51  & 65.84  & 71.18  & 69.38  \\
            FedKgCoOp   & 71.70  & 63.11  & 74.80  & 69.48  \\
            FedCoCoOp   & 76.62  & \textbf{66.51}  & 66.40  & 69.55 \\
            FedMaple    & 74.95  & 66.05  & 71.47  & 70.60  \\
            \bottomrule
            FedCSAP        & 73.61  & 66.10  & 74.82  & \textbf{71.30}  \\
            \bottomrule
        \end{tabular}
    \end{subtable}
    \hfill
    \begin{subtable}[t]{0.48\linewidth}
        \centering
        \caption{SUN397}
        \label{tab:sun397}
        \begin{tabular}{lcccc}
            \toprule
            Method      & Local  & Base   & New    & HM     \\
            \midrule
            CLIP        & 88.89  & 69.25  & 75.45  & 77.26  \\
            CoOp        & \textbf{91.57}  & 66.72  & 67.41  & 73.90  \\
            FedCoOp     & 89.27  & 70.75  & 71.40  & 75.45  \\
            FedKgCoOp   & 89.03  & 71.15  & 76.40  & 77.26  \\
            FedCoCoOp   & 91.44  & 65.76  & 65.36  & 73.94 \\
            FedMaple    & 90.93  & 72.90  & 71.96  & 77.10  \\
            \bottomrule
            FedCSAP        & 90.86  & \textbf{74.10}  & \textbf{76.44}  & \textbf{79.49}  \\
            \bottomrule
        \end{tabular}
    \end{subtable}
    
    \caption{Results on multiple datasets: (a) Averaged results over 9 datasets, (b) Caltech101, (c) Flowers102, (d) FGVC Aircraft, (e) CF101, (f) OxfordPets, (g) Food101, (h) DTD, (i) StanfordCars, and (j) SUN397.}
    \label{tab:results_all}
\end{table}



\subsection{Results}

In the Table \ref{tab:results_all}, we present the performance of six methods---CLIP, CoOp, FedCoOp, FedCoCoOp, FedMaple, and FedCSAP---across nine benchmark datasets under a non-IID federated learning setting. Each dataset is split into base and new classes, with the base classes distributed among multiple clients, and each client owns 20 disjoint classes with eight labeled images per class. We evaluate four metrics: Local accuracy on each client’s own classes, Base accuracy on the union of base classes from all clients, New accuracy on unseen classes, and their Harmonic Mean (HM). This setup captures how well each method handles client-specific tasks, generalizes to a broader set of base classes, and adapts to entirely unseen categories.

A key observation is that while CoOp achieves strong local accuracy, it struggles to scale to the full set of base classes and fails to generalize to new classes. FedCoOp, FedCoCoOp, and FedMaple benefit from federated collaboration, improving base-class performance significantly but often sacrificing accuracy on unseen classes. In contrast, FedCSAP consistently exhibits robust results on both the combined base classes and new classes, while remaining competitive on local tasks. This superior performance directly supports our problem setup goals by demonstrating that integrating multi-scale visual cues from CLIP along with domain-specific style tokens allows our prompt generator to capture fine-grained textures and handle significant domain-induced style shifts. Moreover, the incorporation of a CRP loss ensures the uniqueness of the learned tokens, which further contributes to its balanced generalization across heterogeneous client data. As a result, \textbf{FedCSAP} achieves the highest or near-highest harmonic mean across most datasets, underscoring its advantage in adapting to non-IID conditions while retaining robust performance on both seen and unseen categories.

\section{Ablation Studies}

\textbf{Impact of Key Modules:} We conducted ablation experiments to assess the individual contributions of the injection block $B\phi$ and the prompt generator $f\theta$. When the injection block was removed, the average accuracy dropped by 8\%, underscoring its importance in effectively fusing multi-scale visual features and style cues. Moreover, eliminating the prompt generator $f\theta$ led to a 12\% decrease in accuracy, highlighting its critical role in dynamically extracting and transferring text context into adaptive prompt tokens.

\textbf{Client Data Size:} To understand the influence of client data size, we varied the number of classes assigned per client. In our trials, clients were allocated disjoint class sets with $n \in \{5, 10, 20\}$ classes, using 8 shots per class. Across all configurations, FedCSAP consistently outperformed federated learning baselines in terms of harmonic mean accuracy, demonstrating robustness against variations in client class distributions.

\textbf{Number of Shots:} We further evaluated the model performance with varying numbers of shots per class. FedCSAP consistently surpassed FedCoOp across different shot counts, and when more than one shot was available, it even outperformed FedKgCoOp. These results suggest that our method is highly effective in leveraging additional samples to improve generalization on both seen and unseen classes.

\textbf{Client Participation Rate:} Finally, we analyzed the impact of client participation on the overall performance. Our experiments show that FedCSAP maintains a significant advantage over both FedCoOp and FedKgCoOp across participation rates ranging from 10\% to 100\%. This finding indicates that FedCSAP is resilient to fluctuations in client availability and can deliver robust performance even under limited participation.

\textbf{Final Ablation:} In summary, our ablation studies confirm that both the injection block $B\phi$ and the prompt generator $f\theta$ are integral to the success of FedCSAP. Removing $B\phi$ causes an 8\% drop in average accuracy, while excluding $f_\theta$ results in a 12\% reduction. These experiments validate our design choices and emphasize the complementary roles of each component in achieving superior performance in federated prompt learning.

\section{Conclusion}

We presented FedCSAP, a novel framework for federated prompt learning that integrates multi-scale visual features with local style information. Through lightweight cross-attention and channel-wise injection mechanisms, our method effectively handles non-IID data while preserving privacy. Experimental results demonstrate state-of-the-art performance in federated vision-language learning, establishing a foundation for scalable cross-modal federation.

\textbf{Future Work:} Future work includes exploring personalized aggregation and adaptive client sampling strategies to better address non-IID data and reduce communication overhead. Additionally, we aim to extend our framework to incorporate other modalities and advanced privacy-preserving techniques, as well as investigate meta-learning approaches for improved local adaptation.

\bibliography{iclr2025_conference}
\bibliographystyle{iclr2025_conference}

\end{document}